\documentclass[twocolumn,twoside]{article}
\usepackage{Arxiv,multicol,times}

\usepackage{epsfig}
\usepackage{graphicx}
\usepackage{hyperref}
\usepackage{amssymb}
\usepackage{algorithm}
\usepackage{algorithmic}
\usepackage{inputenc}
\usepackage{amssymb}
\usepackage{amsmath}
\usepackage[overload]{empheq}
\usepackage{hyphenat}
\usepackage{float}
\usepackage{placeins}

\DeclareMathOperator*{\argmax}{argmax}
\usepackage[pagewise]{lineno}
\emergencystretch=\maxdimen
\PUBYEAR{2019}
\VOL{1}
\newcommand\startpage{1}
\usepackage{fancyhdr}
\usepackage{ifthen}
\usepackage{lastpage}
\fancypagestyle{mypagestyle}{%
    \fancyhf{} 
    \fancyhead[RO]{Boudraa, Hidouci and Michelucci}
    \fancyhead[LO]{\thepage}
    \fancyhead[LE]{Degraded Historical Documents Images Binarization Using a Combination of Enhanced Techniques}
    \fancyhead[RE]{\thepage }
    \renewcommand{\headrulewidth}{0pt}
}
\begin{document}
\fancypagestyle{plain}{%
  \renewcommand{\headrulewidth}{0pt}%
  \fancyhf{}%
  \fancyfoot[R]{1
  }
}
\sloppy

\title{
Degraded Historical Documents Images Binarization Using a Combination of Enhanced Techniques}

\author{{\bf Omar Boudraa$^1$, Walid Khaled Hidouci$^1$ and Dominique Michelucci$^2$} \\[1em]
$^1$Laboratoire de la Communication dans les Syst\`emes Informatiques, Ecole nationale Sup\'erieure d'Informatique (ESI),\\
BP 68M, 16309, Oued-Smar, Alger, Alg\'erie. http://www.esi.dz\\
\textit{\{o\_boudraa, w\_hidouci\}@esi.dz} \\[1em]
$^2$Laboratoire LIB, Universit\'e de Bourgogne,\\
BP 47870, 21078, DIJON CEDEX, France\\
\textit{Dominique.Michelucci@u-bourgogne.fr}}

\maketitle

\begin{abstract} 
	Document image binarization is the initial step and a crucial in many document analysis and recognition scheme. In fact, it is still a relevant research subject and a fundamental challenge due to its importance and influence. This paper provides an original multi-phases system that hybridizes various efficient image thresholding methods in order to get the best binarization output. First, to improve contrast in particularly defective images, the application of CLAHE algorithm is suggested and justified. We then use a cooperative technique to segment image into two separated classes. At the end, a special transformation is applied for the purpose of removing scattered noise and of correcting characters forms. Experimentations demonstrate the precision and the robustness of our framework applied on historical degraded documents images within three benchmarks compared to other noted methods.
\end{abstract}


\section{Introduction}
\label{intro}
As a proof of the inestimable information in historical documents, they always remain an important cultural and scientific reference for information retrieval process. Document image binarization is one of the primordial and critical steps in most document analysis systems \cite{Ref1}. For example, this process can facilitate the segmentation of the document which may improve the result quality of many OCR (Optical Character Recognition) systems.

Binarization or Thresholding is an operation that allows separation between the pertinent objects and the background of the document image, where an object is any pattern inscribed on the document (characters, graphemes, words, graphics, etc). The thresholding principle is to get a bitonal image (black and white) from a grayscale image, which considerably reduces the amount of data to be treated. However, common difficulties for image binarization methods are to deal with degraded documents containing non-textual data, such as: smear, shadow, graphics, blur or noise, and other artifacts including: non-uniform illumination, background variation and faint characters.

The proposed algorithm is based on three processing steps, namely, \textit{Preprocessing, Hybrid binarization} and \textit{Post-processing}. In preprocessing, weak contrast is enhanced using CLAHE (\textit{Contrast Limited Adaptive Histogram Equalization}) method \cite{Ref2}. The binarization step involves a hybridization between three well-known thresholding algorithms (Otsu's \cite{Ref3}, Multilevel Otsu's \cite{Ref4} and Nick's Methods \cite{Ref5}) while local average contrast value represents the decision criterion. In the final step, post-processing comprises noise removal by connected components analysis and morphological operations which aim to improve the final binarized image \cite{Ref6}.

This paper is organized as follows:  the next section  presents a review of related works devoted to document image binarization problem. Section~\ref{sec:3} provides a detailed description of our proposed algorithm. Section~\ref{sec:4} is consecrated to experimentations, results, some analysis and the comparison of our contribution with other existing methods. Finally, a critical discussion concludes the paper in Section~\ref{sec:5}.

\section{Related Work}
\label{sec:2}
In recent years, various techniques of document image binarization are used and reported. Mainly, the proposed methods may be classified into the following:
\begin{enumerate}
	\item \textit{Global Thresholding Methods.}
	\item \textit{Adaptive Thresholding Methods.}
	\item \textit{Hybrid Thresholding Methods.}
\end{enumerate}

\subsection{Global Thresholding Methods}

Global thresholding methods use the same threshold value for the entire image. In this case, the intensity values are compared with the same threshold. The threshold value is calculated on the basis of the complete input image.

\begin{figure}[H]
	\centering
	\includegraphics[width=.4\textwidth]{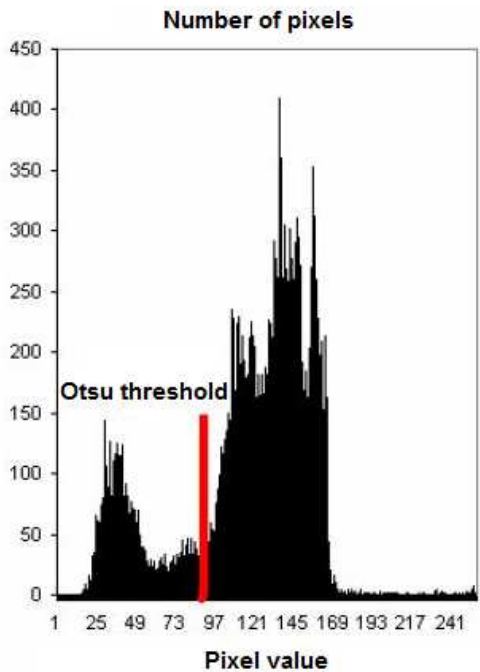}
	\caption{Otsu's thresholding by Histogram analysis.}
	\label{fig:1}       
\end{figure}

The majority of the global techniques uses the histogram values for the computation of the threshold to be applied. Otsu's method \cite{Ref3} is the most common and one of the best global thresholding methods (see Fig.~\ref{fig:1}). This algorithm assumes that an image follows a bimodal histogram and tries to classify the image pixels into foreground (text) or background (non-text) pixels. Then it calculates the optimal threshold separating those two classes so that the intra-class variance is minimal (here, $i$ refers to the class number or index). Otsu demonstrates that minimizing the intra-class variance $v_{i}$ is equivalent to maximize the inter-class variance $v_{b}$, which can be represented with the following equation:

\begin{equation}
T = \argmax_t [v_{b} = |v_{0} - v_{1}| = \omega_{0}(t) \omega_{1}(t)(\mu_{0}(t)-\mu_{1}(t))^{2}]
\end{equation}

\noindent where, for a given threshold $t$, $v_{0,1}$ is the intra-class variance, $\omega_{0,1}(t)$ is the class weight (probability) while  $\mu_{0,1}(t)$ is the class mean value. The algorithm stops after running through the whole range of threshold values $t \in [0,255]$ and keeps the value that maximizes the inter-class variance.

\paragraph{Remark 1.} This approach is easy and appropriate for simple and good quality documents; however, it does not work well for non-uniform background or noisy documents.

\subsection{Adaptive Thresholding Methods}

On the other side, adaptive (or local) thresholding methods compute a threshold for each pixel (or set of pixels) in the image which depends on the content of its neighborhood.

Bernsen \cite{Ref7} first proposed an adaptive thresholding approach based on image contrast. Its method computes the local threshold value by using the mean value of the maximum ($max(i,j)$) and minimum ($min(i,j)$) intensities inside a local neighborhood window centered at pixel $(i,j)$. However, if the contrast $C(i,j) = max(i,j) - min(i,j)$ is smaller than a certain contrast threshold (e.g. $k$ = 15), the pixels inside the window may be set to the most appropriate class (background or  foreground). Nevertheless, this method works well only for big contrast values. 

Niblack's algorithm \cite{Ref8} calculates a pixel-wise threshold which is based on the local mean $m$ and the standard deviation $s$ of all the gray level pixels in a sliding rectangular window, its threshold $T$ is defined as:

\begin{equation}
T = m + k \times s
\end{equation}

\noindent where $k$ is a constant used to adjust the objects boundaries. Although the Niblack's method correctly identifies text areas of the document, it produces a great amount of background noise.

Sauvola and Pietikainen \cite{Ref9} proposed an improvement on the method proposed by Niblack which aims to overcome its background noise problem. Their threshold $T$ is given by the formula (\ref{eq:3}) below:

\begin{equation}
T = m \times \left(1 - k  \times \left(1 - \frac{s}{R} \right) \right)
\label{eq:3}
\end{equation}

The usual values of $k$ = 0.5 and $R$ =128 are recommended. This method outperforms Niblack's algorithm but produces often thin and broken characters. 

\begin{figure*}
	\centering
	\includegraphics[width=\textwidth]{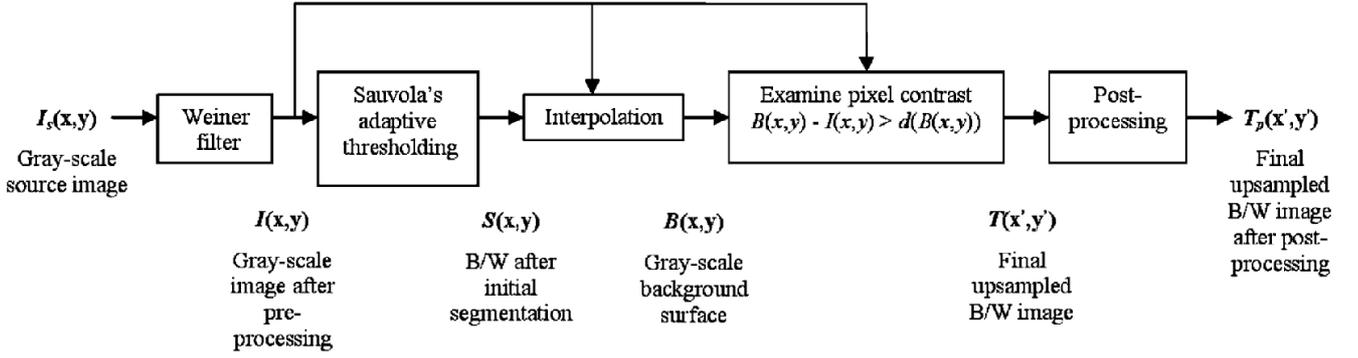}
	\caption{Diagram of Gatos's methodology for degraded historical document image binarization \cite{Ref10}.}
	\label{fig:2}       
\end{figure*}

Gatos et al. \cite{Ref10} proposed a multi-stage document image binarization method. The first step consists in applying a low pass Wiener filter to reduce noise and correct the image contrast; it is followed by using Sauvola’s method to provide preliminary image segmentation. Then, the background area is estimated via intensity analysis technique. In the last step, both images are used to generate the final thresholding result (see Fig.~\ref{fig:2}). The main drawback of this technique is it allows only the enhancement of the textual part of document.

Khurshid et al. \cite{Ref5} proposed an enhanced variant of Niblack's method that they called Nick's method. Their threshold is calculated as follows:

\begin{equation}
T = m + k \times \sqrt{\frac{\left( \sum\limits_{i=1}^{\mathit{NP}} (p_{i}^{2}-m^{2}) \right)}{\mathit{NP}}} 
\end{equation}

where, \begin{itemize}
	\item $k$ is the Nick's factor having value between $-0.2$ and $-0.1$. The smaller the value of $k$, the thicker the binarized stroke, and the noise in resulting image.
	\item $p_i$ is the pixel value in the grayscale image.
	\item $\mathit{NP}$ is the number of pixels in the sliding window.
	\item $m$ is the mean value.
\end{itemize}

Recently, Khan and Mollah \cite{Ref11} have presented a novel algorithm which is based on three phases. First the preprocessing aims to eliminate noise and enhance the document image quality. Second, a variant of Sauvola's Binarization method is applied to this image. Finally, the post-processing analyses small connected components and removes some of them.

\paragraph{Remark 2.} This category of methods generally performs better for degraded and low quality images. However, it has some drawbacks as the dependence on the windows size and the excessive computation time. Finally they often generate bad results in presence of back to front interference problems.

\subsection{Hybrid Thresholding Methods}

As we have seen previously, global and adaptive thresholding approaches have both advantages and disadvantages. Researchers have tried to take benefits from the strengths of both approaches. This gave rise to what is called the hybrid binarization approach. It combines the advantages of both solutions: speed and whole image consideration of global thresholding, flexibility and efficiency on foreground information extraction of adaptive thresholding.

\begin{figure}[H]
	\centering
	\includegraphics[width=.55\textwidth]{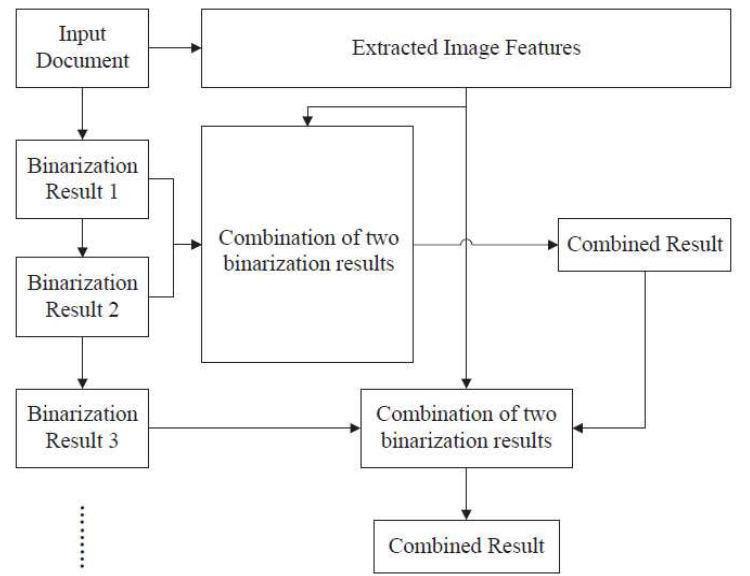}
	\caption{Flowchart of document image binarization combination \cite{Ref12}.}
	\label{fig:3}       
\end{figure}

Su et al. \cite{Ref12} presented a new algorithm that combines different thresholding methods (Otsu's and Sauvola's methods; Gatos's and Su's methods; Lu's and Su's methods) with the aim to improve the document image binarization quality. The proposed algorithm begins with contrast and intensity features extraction which make easier the separation between the foreground text and document background. Then, for a given document image, different binarization methods are used to create many binarized images (see Fig.~\ref{fig:3}). Hence, the document image pixels are classified into three sets, namely, foreground pixels, background pixels and uncertain pixels according to the following formula:

\begin{equation}
P(x,y)=
\begin{cases}
0, & \text{if~}  \sum\limits_{i=1}^{n} B_{i}(x,y) = 0 \\
1, & \text{if~}  \sum\limits_{i=1}^{n} B_{i}(x,y) = n \\
uncertain, & \text{Otherwise} 
\end{cases}
\end{equation}

\noindent where $(x,y)$ are the Cartesian coordinates of the pixel, $n$ is the number of participating methods and $B_i(x,y)$ denotes the binarization result of pixels $P(x,y)$ generated by the $i^{th}$ binarization method. Finally, the uncertain pixels are inspected and set to background or foreground in accordance with their distance to local background and foreground pixels. However, the main drawback of this method is when both methods miss-classify conjointly foreground or background pixels.

In order to achieve better quality of binarization, Sokratis et al. \cite{Ref13} presented a hybrid approach which tries to combine both global and local thresholding methods. The proposed algorithm is summarized as:

\begin{itemize}
	\item Application of Global Algorithm to the entire image (\textit{Iterative Global Thresholding} IGT).
	\item Noisy area detection.
	\item Application of IGT technique locally to each detected area.
\end{itemize} 

For IGT technique, an iterative procedure is executed which involves subtraction of current threshold $T_i$ from each pixel and histogram equalization. The full procedure is repeated until the absolute difference between $T_i$ and $T_{i-1}$ is strictly less than some value (e.g. 0.05).

As for noisy areas detection, the image is first segmented. Then, the amount of black pixels in each segment is estimated. Segments which contain more black pixels on average compared to others are selected when:

\begin{equation}
f(S) > m + k \times s
\label{eq:6}
\end{equation}

noindent where $f(S)$, $m$ and $s$ represent respectively the frequency of the black pixels in the segment $S$, the mean and the standard deviation of the foreground pixels of the whole image, while parameter $k$ controls the sensitivity of the detection areas in this method.

Moghaddam et al. proposed a novel hybrid binarization method which implicates a combination between the Ensemble-of-Expert (EoE) framework \cite{Ref14} and the Grid-based Sauvola's technique as an entry \cite{Ref15}. The EoE framework joins the outputs of various binarization techniques. Then, it produces their confidence maps. The EoE framework recognizes the best experts for a given document image, and analyses their results to generate the final document image. The Grid-based Sauvola's technique has three parameters to adjust; by making use of the grid based modeling, the allocated resources (runtime and memory) might be considerably reduced. A post-processing based on texture analysis is finally applied to the output document image.

Mitianoudis and Papamarkos exploit a new methodology of improving document image binarization using three-stage algorithm. First, the Background is removed using an iterative median filter. Then, the miss-classified pixels are separated by a combination between Local Co-occurrence Mapping (LCM) and Gaussian Mixture clustering. Finally, isolated components are identified and suppressed using a morphology operator \cite{Ref16}.

\section{Document Image Binarization Approach}
\label{sec:3}
Our method is based on hybrid thresholding approach using three performing binarization techniques, joined with preprocessing and post-processing steps which are used to correct and improve the obtained results.
\begin{figure}[H]
	\centering
	\includegraphics[width=.5\textwidth]{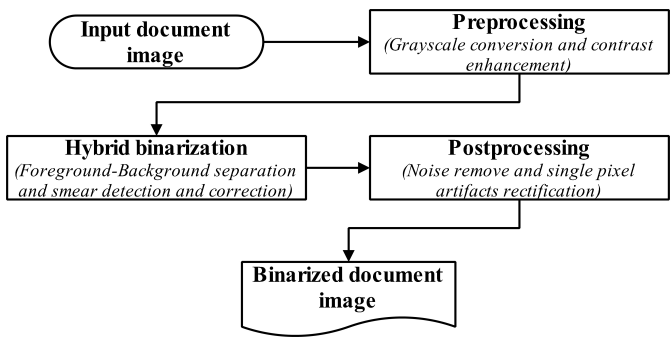}
	\caption{Diagram of proposed system for historical document image binarization.}
	\label{fig:4}       
\end{figure}
\subsection{Motivations of the Proposed Approach}

Compared to other methods, our approach has its proper advantages. Here, we propose to take advantage of CLAHE method which is used to improve contrast in images while the over-amplification of noise is prevented \cite{Ref2}. In addition, the cooperation between various algorithms may improve the binarization quality and may reduce considerably the execution time compared to launching adaptive thresholding methods only. Post-processing is finally applied to eliminate small noisy elements and correct characters format. In fact, preprocessing and post-processing stages have a major impact on our document binarization process. A diagram of the proposed approach is presented in Fig.~\ref{fig:4}.

\subsection{Proposed Algorithm}

The full proposed document image binarization algorithm steps are explained and detailed as follows: 

\subsubsection{Preprocessing} 

Preprocessing consists in eliminating the defects associated with the input document image in order to facilitate the hybrid binarization step.

\begin{figure*}
	\centering
	\includegraphics[width=\textwidth]{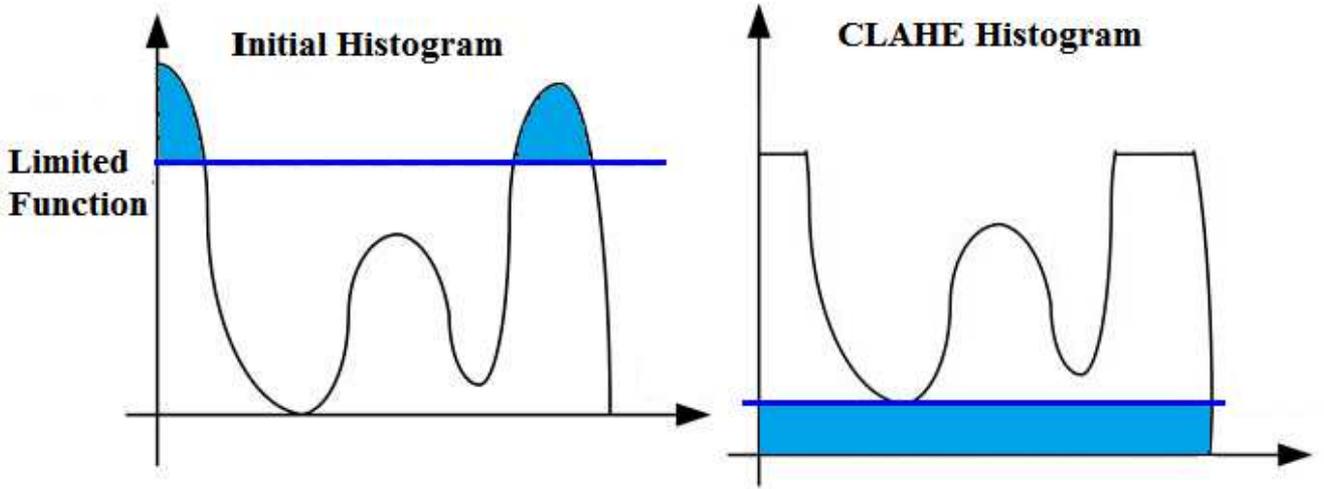}
	\caption{Contrast Limited Adaptive Histogram Equalization (CLAHE) and Limited Function.}
	\label{fig:5}       
\end{figure*}

First, a color document must be converted to grayscale. Then, image contrast is enhanced using CLAHE method, because it is more favorable for improving the local contrast in each section (tile) of the image (see Fig.~\ref{fig:5}). In addition, it limits the over-amplification of noise included in homogeneous regions compared to other adaptive histogram equalization algorithms \cite{Ref2}.

Nevertheless, it is not a good idea to directly apply the CLAHE method to all images; because, in some cases, it can deform the objects edges and increase noise,  altering relevant information. So, we propose to make use of contrast value as decision criterion, if this value is below a certain threshold $T_{ctr}$ (i.e. weak contrast) the CLAHE is therefore performed. Many definitions of contrast exist. Here, we opt for \textit{Michelson's contrast} formula \cite{Ref6}, as:

\begin{equation}
Ctr =  \frac{I_{max} - I_{min}}{I_{max} + I_{min} + \epsilon}
\end{equation}

\noindent where, $I_{max}$ and $I_{min}$ are respectively the highest and lowest intensity values of the document image, $\epsilon$ is a tiny positive value, which is considered only if the highest intensity is zero value. As the global contrast is often not significant, we propose to calculate the local contrast by estimating the average value of all local contrasts in sliding a window of size $3 \times 3$. We found $T_{ctr}$ = 0.02 to be good choice. 
\subsubsection{Hybrid Binarization}

In this second step, we proceed to hybrid thresholding by combining the binarization issues of three common methods (namely: Otsu's method \cite{Ref3}, Nick's method \cite{Ref5} and Multi Level Otsu's method as proposed in \cite{Ref4}).

Although Otsu's method remains an attractive choice, TSMO method (Two-Stage Multi-threshold Otsu's method) outperforms Otsu's method in particular cases, especially when there are more than two different classes \cite{Ref4}. However, Nick's method deals better in presence of background and foreground intensities variation.

\begin{figure}[H]
	\centering
	\includegraphics[width=0.5\textwidth]{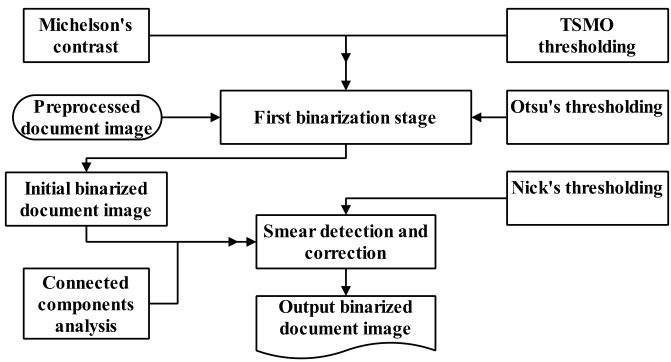}
	\caption{Flowchart of proposed hybrid thresholding stage.}
	\label{fig:6}       
\end{figure}

\begin{figure*}
	\centering
	\begin{tabular}{c}
		\includegraphics[width=.7\textwidth]{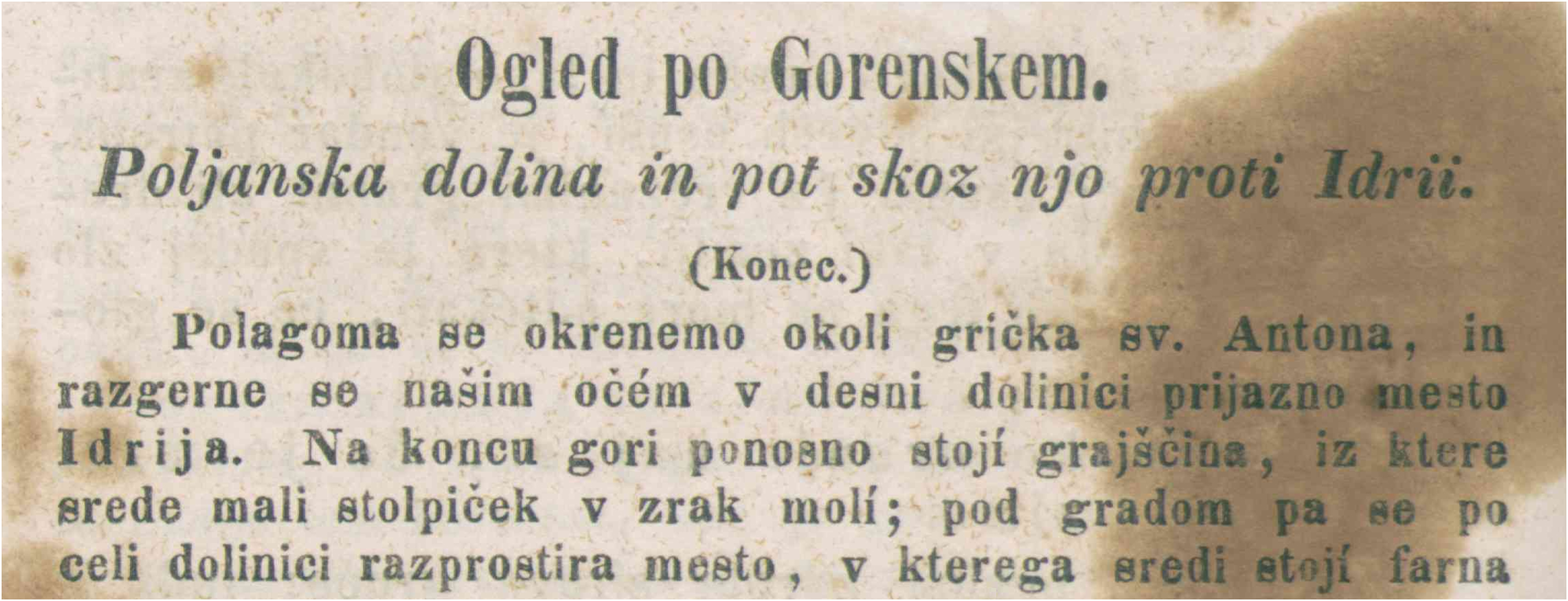}   \\
		(a)\\
		\includegraphics[width=.7\textwidth]{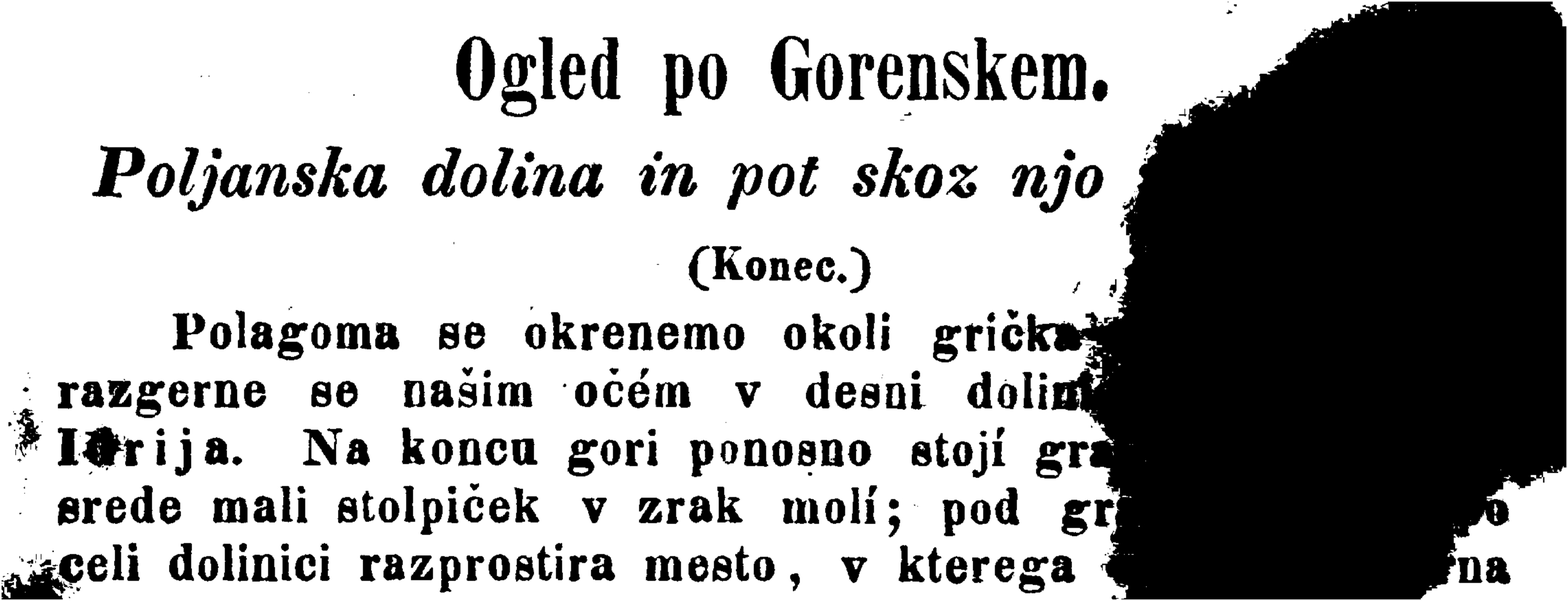}    \\
		(b)\\
		\includegraphics[width=.7\textwidth]{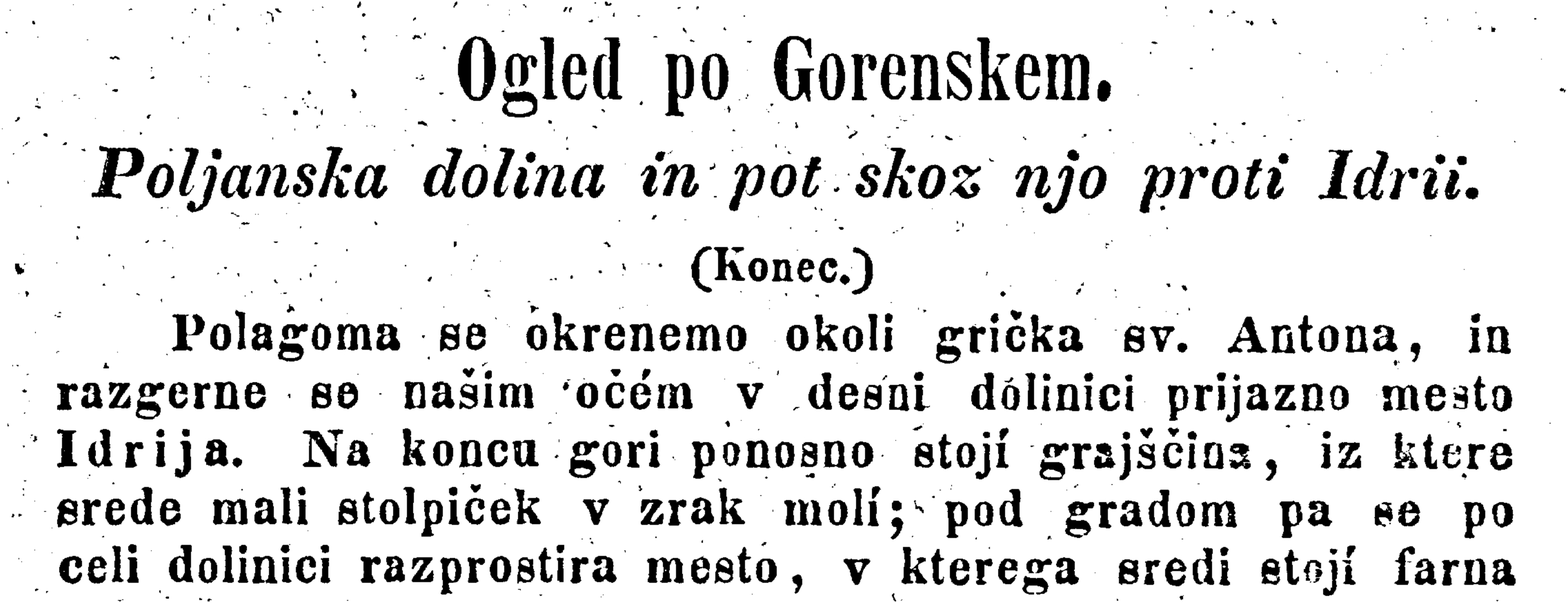}    \\
		(c)\\
		\includegraphics[width=.7\textwidth]{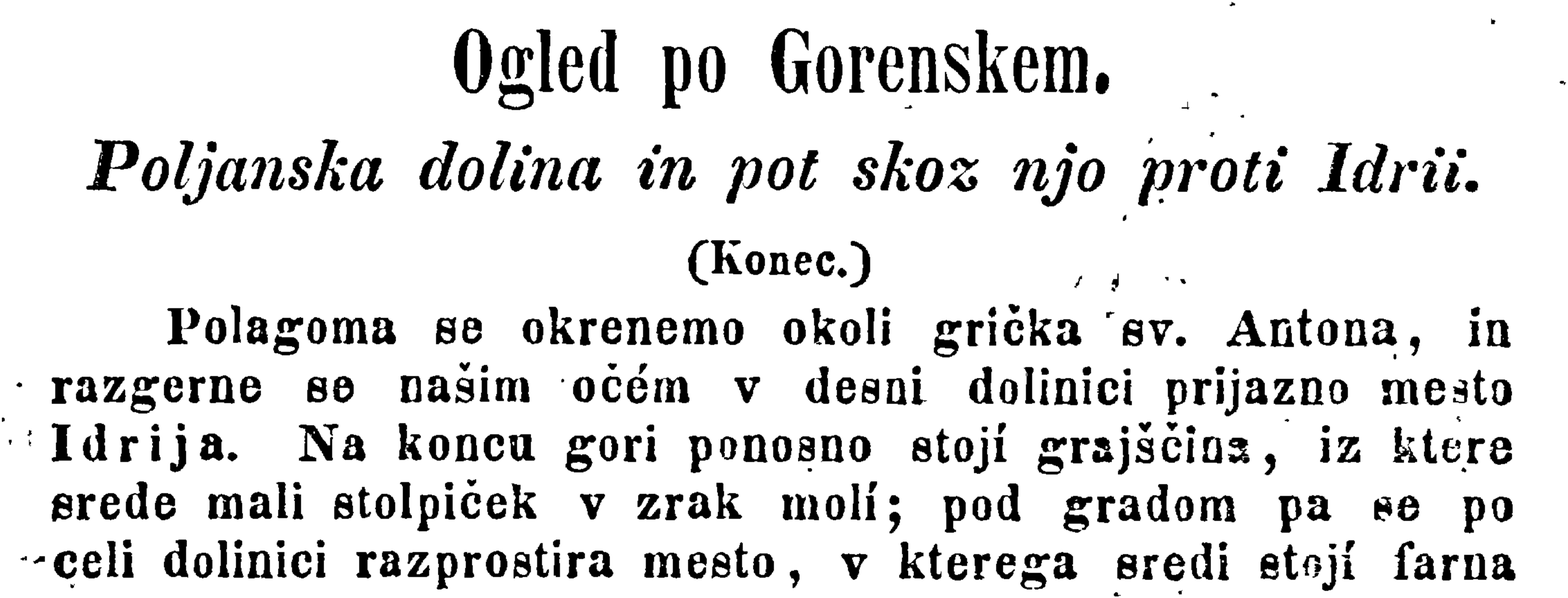}    \\
		(d)\\
		
	\end{tabular}
	\caption{Smear detection and correction \cite{Ref17} (a) Original image (b) Otsu (c) Nick (d) Our hybrid method.}
	\label{fig:7}
\end{figure*}

In addition, we suggest again to use contrast as a decision criterion (see Fig.~\ref{fig:6}). Thus, according to registered contrast value, we divide all document images qualities into four categories of contrast according to three thresholds ($T_1$ , $T_2$ and $T_3$), for each category we apply a specialized binarization method as follows: 

\begin{itemize}
	\item \textbf{Low-contrast image ($Ctr \leqslant T_1$):}
	
	We binarize document image using the second detected $T_{O2}$ threshold of TSMO method.	
	
	\item \textbf{Fuzzy-contrast image ($T_1 < Ctr \leqslant T_2$):} 
	
	This ambiguous interval involves additional verifications in order to select the most adequate technique to use. Therefore, we proceed as follows:
	
	\textbf{If} ($D_{min} \leqslant |T_{O2} - T_O| \leqslant D_{max}$ \textbf{and NB} ($T_O < Pixels~intensities \leqslant T_{O2}$) $\leqslant P \times$ \textbf{NB} ($Pixels~intensities \leqslant T_O)$) \textbf{Then} we perform binarization using the second detected $T_{O2}$ threshold of TSMO method. $P$ is a user factor comprised between 0 and 1. 
	
	\textbf{Else} we perform binarization using Otsu's method (with $T_{O}$ threshold).
	
	\item \textbf{Medium-contrast image ($T_2 < Ctr \leqslant T_3$):}
	
	We apply simple Otsu's method.
	
	\item \textbf{High-contrast image ($Ctr > T_3$):} 
	
	We binarize document image using the smallest threshold $T_{O1}$ of TSMO method.	
\end{itemize}
In the second part of this stage, big agglomerations of pixels are then processed by analyzing the connected components. Instead, the method proposed by Sokratis et al. \cite{Ref13} divides the document image into segments, in order to detect malicious objects that are most likely to contain unwanted pixels, and uses  the same definition for background noise, given in Eq. (\ref{eq:6}). Here, and after having detected these malicious objects (smear), we proceed to a second binarization using a local thresholding algorithm, namely: Nick's method (see Fig.~\ref{fig:7}). 

\subsubsection{Post-processing}

In the last step, a series of post-processing operations  eliminate noise and improve the quality of text regions by filling various gaps and holes, and removing unwanted connected components (see Alg.~\ref{alg:1}). 

\begin{algorithm}[H]
	\caption{Post-processing of a document image}
	\begin{algorithmic}[1]
		\REQUIRE One binarized document image.
		\ENSURE Gaps filling and small isolated connected components filtering of the input binarized document image.	
		\STATE Remove foreground pixels that do not connect with others;
		\STATE Fill one-pixel gaps of the entire image;
		\STATE Find out all the connected components;
		\STATE Calculate the mean $m$ and the standard deviation $s$ of the amount of pixels included in each connected component;	
		\FORALL {CC $\in$ the document image}
		\IF{Pixels number of current CC $>$ ($(\lambda \times m) / s$)} \STATE Its pixels are changed to background;
		\ENDIF
		\ENDFOR
		\STATE Remove pixels with convexity artifacts;
		\STATE Fill pixels with concavity artifacts;		
		\RETURN The final improved document image;
	\end{algorithmic}
	\label{alg:1}
\end{algorithm}

\begin{figure*}
	\centering
	\includegraphics[width=\textwidth]{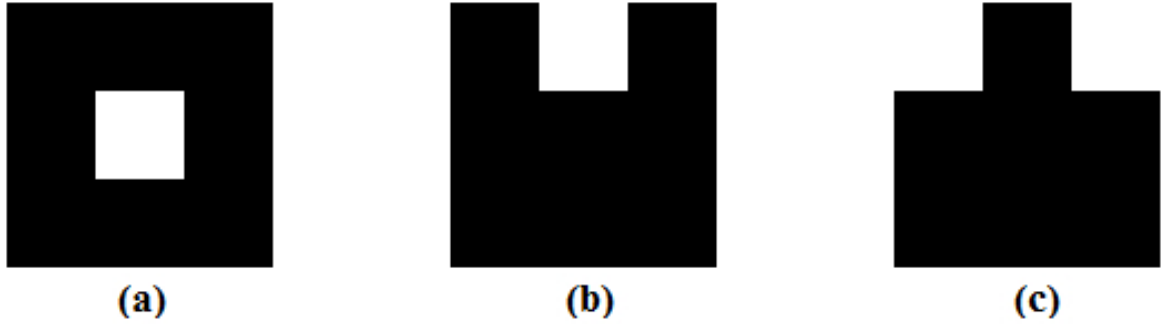}
	\caption{Single pixel artifacts after the document binarization \cite{Ref6} (a) Single pixel gap (b) Single pixel concavity (c) Single pixel convexity.}
	\label{fig:8}       
\end{figure*}

Note that $\lambda$ parameter is set empirically to 15. Instructions $2^{th}$, $10^{th}$ and $11^{th}$ aim to correct the text defects introduced by the document image binarization (namely: single pixel gaps, convexities and concavities). They refine the text stroke edge and its interior region (see Fig.~\ref{fig:8}). These series of morphological filters enhance the binarization of the document image.  This leads to better objects extraction and errors minimization, which helps remarkably in increasing the accuracy of our method.
\section{Experimental Results}
\label{sec:4}
In this section, a summary of performed tests and obtained results is presented to demonstrate the efficiency and the robustness of our image binarization approach. Hence, we test it on a variety of document images (printed and handwritten, net and noisy, good and low-quality, simple and complex) taken from three popular images datasets (see Sec.~\ref{sec:41}). We report the performance of the image binarization techniques in terms of four common evaluation criteria (as detailed in Sec.~\ref{sec:42}).

Then, we compare our approach with the state of the art thresholding methods such as Otsu \cite{Ref3}, Nick \cite{Ref5}, Niblack \cite{Ref8}, Sauvola \cite{Ref9}, Khan \cite{Ref11}, Moghaddam \cite{Ref14,Ref15} and Mitianoudis \cite{Ref16}. We compare with these methods because they are common and widely-used for image binarization problems, and numerous algorithms are based on them. Also, we provide a quick runtime analysis in terms of registered time (in Seconds) of our approach, compared to a global thresholding method (Otsu) and to a local thresholding method (Nick).

\subsection{Datasets}
\label{sec:41}
For examinations, we collected a variety of documents types, such as printed and handwritten, of different languages and scripts, with bad and non-uniform illumination, and added blurr or noise, taken from three well-known datasets.

The first dataset consists of 8 machine-printed and 8 handwritten images of the DIBCO 2013 sample dataset with related ground truth, which were selected from the IMPACT project, the Library of Congress and the TranScriptorium project \cite{Ref17}; they were provided to the participants in order to tune their algorithms in the competition. These images were built so that some particular degradations appear.

The second dataset (H-DIBCO 2014) comprises 10 handwritten document images with the associated ground truth \cite{Ref18}. This collection suffers from some sort of degradations such as low contrast, variable background intensity and faint characters.

The third dataset (H-DIBCO 2016) consists of 10 handwritten document images which have been produced from the ABP and the StAM -- Grimm collections \cite{Ref11}, their ground truth images were constructed manually for the contest. The ABP collection comprises ritual register and index pages while the StAM -- Grimm collection contains principally letters, greeting cards and postcards.

\subsection{Evaluation Criteria}
\label{sec:42}
To measure the document binarization efficiency, obtained results are evaluated and compared using the following measures, as suggested by Pratikakis et al \cite{Ref11}:
\subsubsection{F-Measure (FM)}
this criterion combines both $Precision$ and $Recall$ in one formula by calculating their harmonic mean as follow:

\begin{equation}
\mathit{FM} = \frac{2 \times Recall \times Precision}{Recall + Precision} 
\end{equation}

\subsubsection{Pseudo F-Measure ($\mbox{FM}_{p}$)}

is similar to F-Measure formula, but it makes use of Pseudo-Recall $R_p$ and Pseudo-Precision $P_p$. These two measures are properly modified using a weighted distance between generated Skeleton and the contour of characters in the Ground-Truth ($GT$) image. Moreover, Pseudo-Recall takes into consideration the local stroke width, while Pseudo-Precision expands to the stroke width of the nearest Ground Truth ($GT$) connected component. Finally, Pseudo-Recall and Pseudo-Precision are normalized and clamped in [0,1] and [0,2] respectively \cite{Ref19}.

\subsubsection{Distance Reciprocal Distortion (DRD)}

it quantifies the distortion for all the $N$ modified pixels as follows:

\begin{equation}
\mathit{DRD} = \frac{\sum\limits_{k=1}^{N} \mathit{DRD_{k}}}{\mathit{NUBN}} 
\label{eq:9}
\end{equation}

\noindent where $DRD_{k}$ is the reciprocal distortion distance of the $k^{th}$ modified pixel and $NUBN$ is defined as the number of non-uniform blocks of $8 \times 8$ sizes in the $GT$ image. $DRD_{k}$ is computed by:

\begin{equation}
\mathit{DRD_{k}} = \sum\limits_{i=-2}^{2} \sum\limits_{j=-2}^{2} |B_{k}(x,y) - GT_{k}(i,j)| \times \mathit{W_{NM}}(i,j)
\label{eq:10}
\end{equation}

As shown in Eq. (\ref{eq:10}), $\mathit{DRD_{k}}$ is equal to the weighted sum of the pixels ($\mathit{W_{NM}}(i,j)$) in the block of $5 \times 5$ sizes at the Ground Truth image ($GT_{k}(i,j)$) that do not have the same grayscale value compared to the $k^{th}$ modified pixel centered at $(x,y)$ in the binarized image ($B_{k}(x,y)$) \cite{Ref20}.

\subsubsection{Peak Signal to Noise Ratio (PSNR)}

it is often used as a similarity measure between the binarized image and the ground truth image, The higher the value of $PSNR$, the better the quality of the binarized image, as:

\begin{equation}
\mathit{PSNR} = 10 \times log_{10} \left(\frac{\mathit{MAX_{I}^{2}}}{\mathit{MSE}}\right)  
\end{equation}

\noindent where : $\mathit{MSE} = \frac{\sum\limits_{i=1}^{N} \sum\limits_{j=1}^{M} (B(i,j) - GT(i,j))^{2}} {N \times M}$

\subsubsection{Score}

it is a computational statistic measure which significatly provides a global indication of efficiency of a particular method inside a competitive environment.

In our case, the four presented measures are taken into consideration, the score value of the $i^{th}$ method with reference to the $j^{th}$ dataset when using the $k^{th}$ evaluation criterion is given by:

\begin{equation}
S_{i,j} = \sum\limits_{k=1}^{N} R_i(j,k)
\end{equation}

\noindent where $N$ is the number of evaluation criteria; $R_i(j,k)$ denotes the individual ranking value. The best method has the lowest accumulated score value.

\subsection{Training Set and Parameters Adjustement}
\label{sec:43}
A training set consisting of attentively selected documents images was generated by collecting samples from each dataset. These documents suffer from different types of degradations and they have been used to properly tune our algorithm parameters.

\begin{figure*}
	\centering
	\includegraphics[width=13cm]{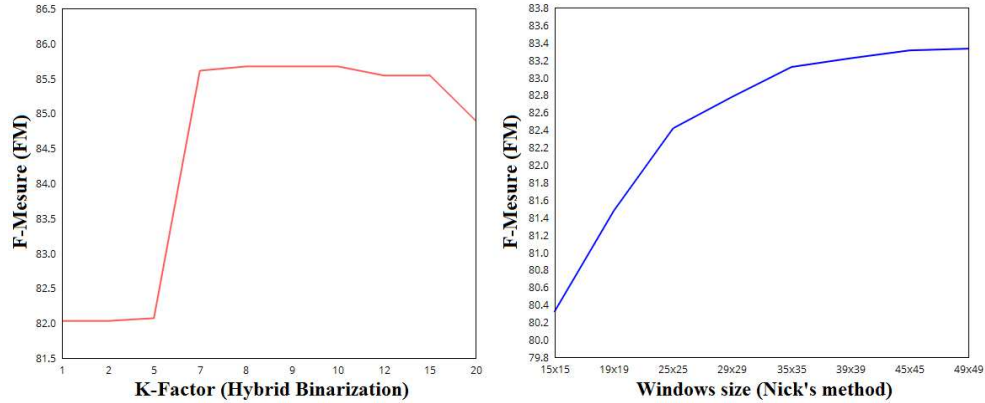}
	\caption{Parameters tuning of hybrid binarization stage.}
	\label{fig:9}       
\end{figure*}

After experimental work, we suggest the following parameter values for the hybrid binarization stage: $T_1=0.03$, $T_2=0.04$, $T_3=0.085$, $D_{min}=5$, $D_{max}=25$, $P=50$, $k=8$. while the $Windows\_size$ of Nick's method is set to $35 \times 35$. Some examples of adjustments are done in Fig.~\ref{fig:9}.

Some of them are fixed by the proposed authors. They identified as: $k=-0.2$ and $Windows\_size=25\times25$ concerning Niblack's method \cite{Ref8}, $k=0.5$, $R=128$ and $Windows\_size=15\times15$ for Sauvola's method \cite{Ref9}.
From the two curves above, on one hand, we can notice that the best value of F-Measure is the one obtained with K-Factor equal to 8. On the other hand, we observe that when the windows size exceed  $35 \times 35$ the improvement becomes not significant; contrariwise, the response time increases considerably.

\subsection{Results}
\label{sec:44}
Table~\ref{tab:1} presents the results of binarization using the three implemented methods over the first dataset (DIBCO 2013) as well as the final ranking. The best method results are highlighted in bold. Overall, our method achieved the best results for all four measures, which confirms clearly its high accuracy in dealing with different documents types under various problems.

The results for the second dataset are listed in Table~\ref{tab:2}. This dataset contains a collection of images derived from degraded historical handwritten documents taken from H-DIBCO 2014 dataset. The results show notable amelioration in all four criteria compared to the last obtained values. For the second time, the worst results are those registered by Niblack's algorithm.  

The last dataset to be evaluated is the one provided within H-DIBCO 2016. Its results are listed in Table~\ref{tab:4}. This dataset contained handwritten images with added blurr, noise and back to front interference problems. To a certain level, all methods were affected by the existence of these complications in these documents, as shown by the registered values in this dataset. Nonetheless, our method and Otsu's method remained the least affected. 
\begin{table}[H]
	\centering
	\caption{Detailed evaluation results of six common methods applied to DIBCO 2013.}
	\label{tab:1}       
	\begin{tabular}{lllllll}
		\hline\noalign{\smallskip}
		Rank & Score & Method & $\mathit{FM}(\%)$ & $\mathit{FM}_p(\%)$ & $\mathit{DRD}$ & $\mathit{PSNR}$\\ 
		\noalign{\smallskip}\hline\noalign{\smallskip}
		\textbf{1} &	\textbf{4} &	\textbf{Proposed} &	\textbf{87.23} &	\textbf{93.40} &	\textbf{4.16} &	\textbf{18.35} \\
		2 &	9 &	Sauvola &	85.02 & 89.77 & 7.58 &	16.94  \\
		3 &	11 &	Moghaddam &	84.9 &	87.41 &	17.02 &	8.25  \\
		4 &	17 &	Otsu &	80.04 &	82.82 &	10.98 &	16.63 \\
		5 &	19 &	Nick & 	80.02 &	83.53 &	12.86 &	15.85  \\
		6 &	24 &	Niblack &	34.12 &	38.01 &	 114.40 & 6.12 \\
		\noalign{\smallskip}\hline
	\end{tabular}
\end{table}

\begin{table}[H]
	\centering
	\caption{Detailed evaluation results of six common methods applied to H-DIBCO 2014.}
	\label{tab:2}       
	\begin{tabular}{lllllll}
		\hline\noalign{\smallskip}
		Rank & Score & Method & $\mathit{FM}(\%)$ & $\mathit{FM}_p(\%)$ & $\mathit{DRD}$ & $\mathit{PSNR}$\\ 
		\noalign{\smallskip}\hline\noalign{\smallskip}
		\textbf{1} &	\textbf{4} &	\textbf{Proposed} &	\textbf{92.40} &	\textbf{96.46} &	\textbf{2.22} &	\textbf{19.09} \\
		2 &	8 &	Otsu &	91.63 &	95.50 &	2.64 &	18.71  \\
		3 &	13 &	Mitianoudis &	89.77 &		90.98 &	4.227 &	18.46  \\
		4 &	17 &	Sauvola &	86.83 &	91.8 &	4.896 &	17.63  	\\	
		5 &	18  &	Nick &	87.80 &	90.50 &	4.47 &	17.59 \\
		6 &	24 &	Niblack &	45.49 &	46.03 & 72.95 &	6.72 \\				
		\noalign{\smallskip}\hline
	\end{tabular}
\end{table}

\begin{table}[H]
	\centering
	\caption{Detailed evaluation results of six common methods applied to H-DIBCO 2016.}
	\label{tab:3}       
	\begin{tabular}{lllllll}
		\hline\noalign{\smallskip}
		Rank & Score & Method & $\mathit{FM}(\%)$ & $\mathit{FM}_p(\%)$ & $\mathit{DRD}$ & $\mathit{PSNR}$\\ 
		\noalign{\smallskip}\hline\noalign{\smallskip}
		\textbf{1} &	\textbf{4} &	\textbf{Proposed} &	\textbf{85.08} &	\textbf{89.81} &	\textbf{5.08} &	\textbf{17.47} \\
		\textbf{1} &	\textbf{6} &	\textbf{Otsu} &	\textbf{86.59} &	\textbf{88.67} &	\textbf{5.58} &	\textbf{17.78}  \\
		3 &	13 &	Khan &	84.32 & 85.64 & 6.94 & 16.59 \\
		4 &	15 &	Sauvola &	82.52 &	86.85 &	7.49 &	16.42	  	\\	
		5 &	20 &	Nick &	81.38 &	83.12 &	10.49 &	15.40 \\
		6 &	24 &	Niblack &	38.76 &	38.64 &	118.51 &	6.41 \\				
		\noalign{\smallskip}\hline
	\end{tabular}
\end{table}

\begin{figure*}
	\centering
	\begin{tabular}{c c}
		\includegraphics[width=.4\textwidth]{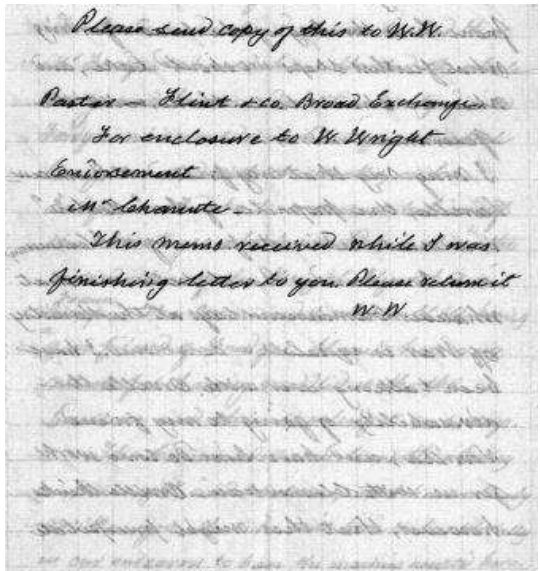}  & 	\includegraphics[width=.4\textwidth]{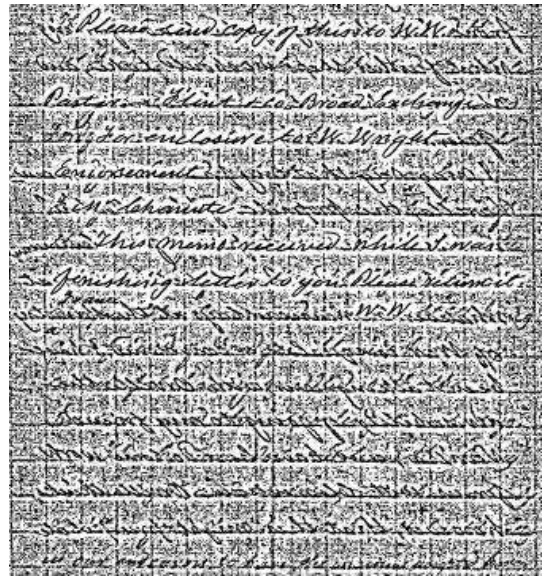}\\
		(a)&(b)\\
		\includegraphics[width=.4\textwidth]{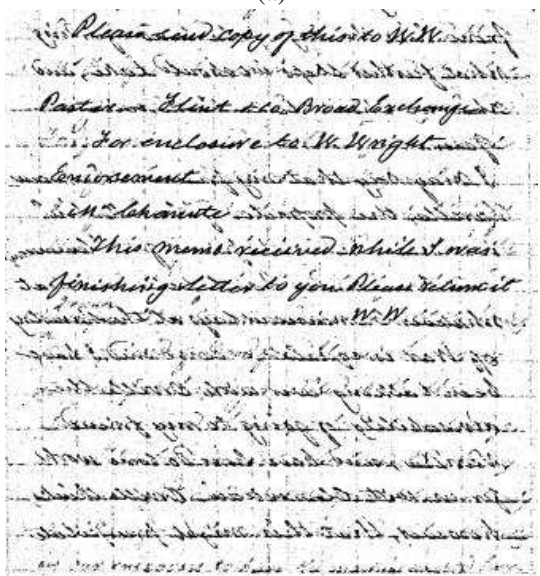}& 	\includegraphics[width=.4\textwidth]{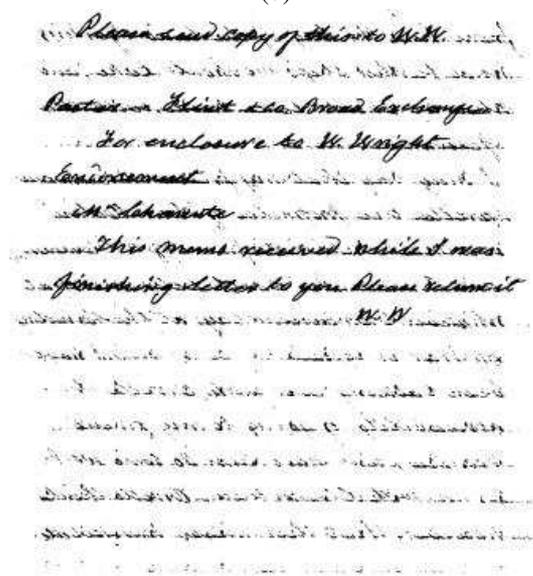}   \\
		(c)&(d)\\
		\multicolumn{2}{c}{\includegraphics[width=.4\textwidth]{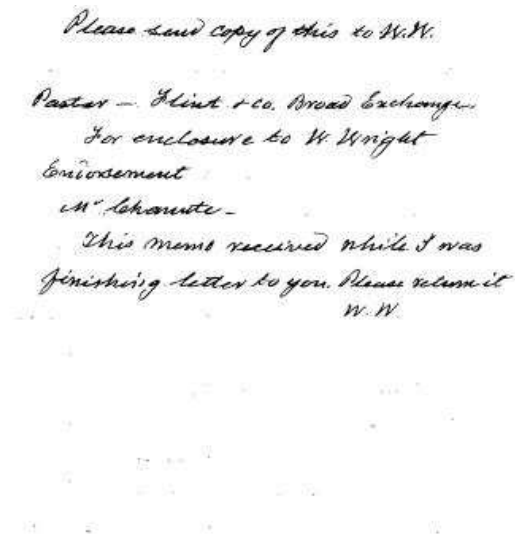}}    \\
		\multicolumn{2}{c}{(e)}\\
	\end{tabular}
	\caption{Binarization results from the three tested algorithms applied to an image taken from DIBCO 2013 dataset \cite{Ref17} (a) Original image (b) Niblack (c) Nick (d) Otsu (e) Proposed method.}
	\label{fig:10}
\end{figure*}

Shown in Fig.~\ref{fig:10} is a binarization results example of the proposed method as well as of Niblack's, Otsu's and Nick's           methods. It is clear that the best separation between foreground and background is done by our proposed method. However, for all other methods, we perceive unevenly the interference between front and back. Hence, the visual and statistical results confirm clearly the high accuracy and the effectiveness of our approach.

\subsection{Computational Time}
\label{sec:45}
The experiments were performed on a 64-bits Windows OS machine with 2.0 GHz AMD Quad-Core Processor and 4 GB of memory. We implemented all algorithms in Open Computer Vision (Intel) integrated in QT/C++ platform without too much optimization effort. Hence, the execution time is not considered as an evaluation criterion. 
\begin{table}[H]
	\centering
	\caption{Average computational time (in Sec) using our proposed method when compared with Otsu's and Nick's methods.}
	\label{tab:4}       
	\begin{tabular}{llll}
		\hline\noalign{\smallskip}
		Method & DIBCO 2013 & H-DIBCO 2014 & H-DIBCO 2016 \\ 
		\noalign{\smallskip}\hline\noalign{\smallskip}
		Otsu &	0.66 &	0.41 &	0.33 \\
		Proposed &	39.40 &	4.09 &	8.33 \\
		Nick &	483.99 & 275.56 & 321.18 \\
		\noalign{\smallskip}\hline
	\end{tabular}
\end{table}

It can be certainly seen from Table~\ref{tab:4} that the runtime is clearly reduced for the three datasets used in experimentations when we apply our proposed method, compared to adaptive Nick's method. The consuming time is reasonably increased when we used global Otsu's method, which is acceptable because of Otsu's method simplicity. Even so, the runtime of our proposed method remains very good taken into account its different stages.

\section{Conclusion and Discussions}
\label{sec:5}
In this paper, we proposed a novel robust approach for image binarization of degraded historical documents. The algorithm is based on hybrid thresholding using three famous binarization methods, combined with preprocessing and post-processing steps to improve binarization quality.

Our experimental results prove the effectiveness and the robustness of this method, and show that it achieves high accuracy in document image binarization on three common datasets containing various types of documents which suffer from different kinds of  problems and defies (background variation, noise presence, low contrast, etc). Nevertheless, our method has had a major inconvenience, namely: the number of algorithm parameters, which is relatively big (eleven). All of them are set apart by long and separate tests.

As a perspective, we plan to use Genetic Fuzzy Trees method as proposed by Ernest et al. \cite{Ref21}. to control the triggering of sub-algorithms, or the values of our software parameters (i.e. generate Fuzzy Rules). Resorting to other methods of Deep Learning represents an other interest idea.

\section*{Acknowledgments}
This research was supported by LCSI and LIB Laboratories. We thank our colleagues from ENSI (Algiers, Algeria) and ESIREM (Dijon, France) who provided insight and expertise that greatly assisted the research, although they may not agree with all of the interpretations of this paper. 

\bibliographystyle{unsrt}
\bibliography{bibliography}

%
%
%
%
%
%

\vspace{5mm}

\noindent{\bf\Large Author Biographies} \vspace{5mm}

\textbf{Omar Boudraa} is a researcher at the LCSI Laboratory of Heigh School of Computer Sciences (ESI), Oued Smar, Algiers, Algeria. His teaching and research interests are in image processing, historical document analysis, networks and systems administration.

\vspace{4mm}

\textbf{Walid Khaled Hidouci} is full professor. Since 2010, he heads the Advanced Data Bases team at LCSI research laboratory. His research interests include: algorithms, database systems, artificial intelligence, parallel and distributed computing and Unix system administration.

\vspace{4mm}

\textbf{Dominique Michelucci} has full professor degree and he is a researcher at Computer Science Laboratory of Burgundy (LIB). His pedagogic and research interests are in image synthesis, geometric transformations and computations of images, modelisation, artificial intelligence, optimization and computer programming.

\end{document}